\documentclass[sigconf]{acmart}
\AtBeginDocument{%
  }

\copyrightyear{2026}
\acmYear{2026}
\setcopyright{cc}
\setcctype{by}
\acmConference[MRAC '26]{4th International Workshop on Multimodal, Generative and Responsible Affective Computing}{November 10--14, 2026}{Rio de Janeiro, Brazil}
\acmBooktitle{4th International Workshop on Multimodal, Generative and Responsible Affective Computing (MRAC '26), November 10--14, 2026, Rio de Janeiro, Brazil}
\acmDOI{10.1145/3840474.3840547}
\acmISBN{979-8-4007-2928-7/2026/11}

\begin{document}

\title{Metric-Aligned Group Relative Policy Optimization for Explainable Deepfake Detection}

\author{Luka Dragar}
\email{ld8435@student.uni-lj.si}
\orcid{0009-0001-0149-3949}
\affiliation{%
  \institution{University of Ljubljana}
  \city{Ljubljana}
  \country{Slovenia}
}

\author{Peter Peer}
\orcid{0000-0001-9744-4035}
\email{peter.peer@fri.uni-lj.si}
\affiliation{%
  \institution{University of Ljubljana}
  \city{Ljubljana}
  \country{Slovenia}
}
\author{Vitomir {\v{S}}truc}
\orcid{0000-0002-3385-5780}
\email{vitomir.struc@fe.uni-lj.si}
\affiliation{%
  \institution{University of Ljubljana}
  \city{Ljubljana}
  \country{Slovenia}
}

\author{Borut Batagelj}
\orcid{0000-0002-8669-2671}
\email{borut.batagelj@fri.uni-lj.si}
\affiliation{%
  \institution{University of Ljubljana}
  \city{Ljubljana}
  \country{Slovenia}
}



\begin{abstract}

Deepfake detectors are commonly evaluated on real/fake classification alone, but real-world applications require explanations of the visual evidence behind each prediction. Joint vision-language models struggle with authenticity assessment, producing ``real'' narratives for manipulated images. To address this, we propose a decoupled predict-then-explain pipeline as our submission to the ACM Multimedia 2026 Explainable Deepfake Detection Challenge~\cite{narang2026explainabledeepfakedetectionchallenge}, with methods trained and evaluated on the XPlainVerse benchmark~\cite{xplainverse2026}. A vision-foundation-model ensemble first classifies the image, then a Qwen3-VL-8B model generates a decision-conditioned explanation of the visual evidence. For fake images, a second Qwen3-VL-8B compresses that text into a short non-expert summary without using the image. We further apply metric-aligned Group Relative Policy Optimization (GRPO) to optimize both explainers against the challenge criteria. Under matched conditions, GRPO raises complex BERTScore-F1 from 0.686 to 0.712 and also improves grounding (entity and evidence F1); our submission ranked first on complex BERTScore-F1 and seventh overall.

\end{abstract}



\begin{teaserfigure}
 \includegraphics[width=\textwidth]{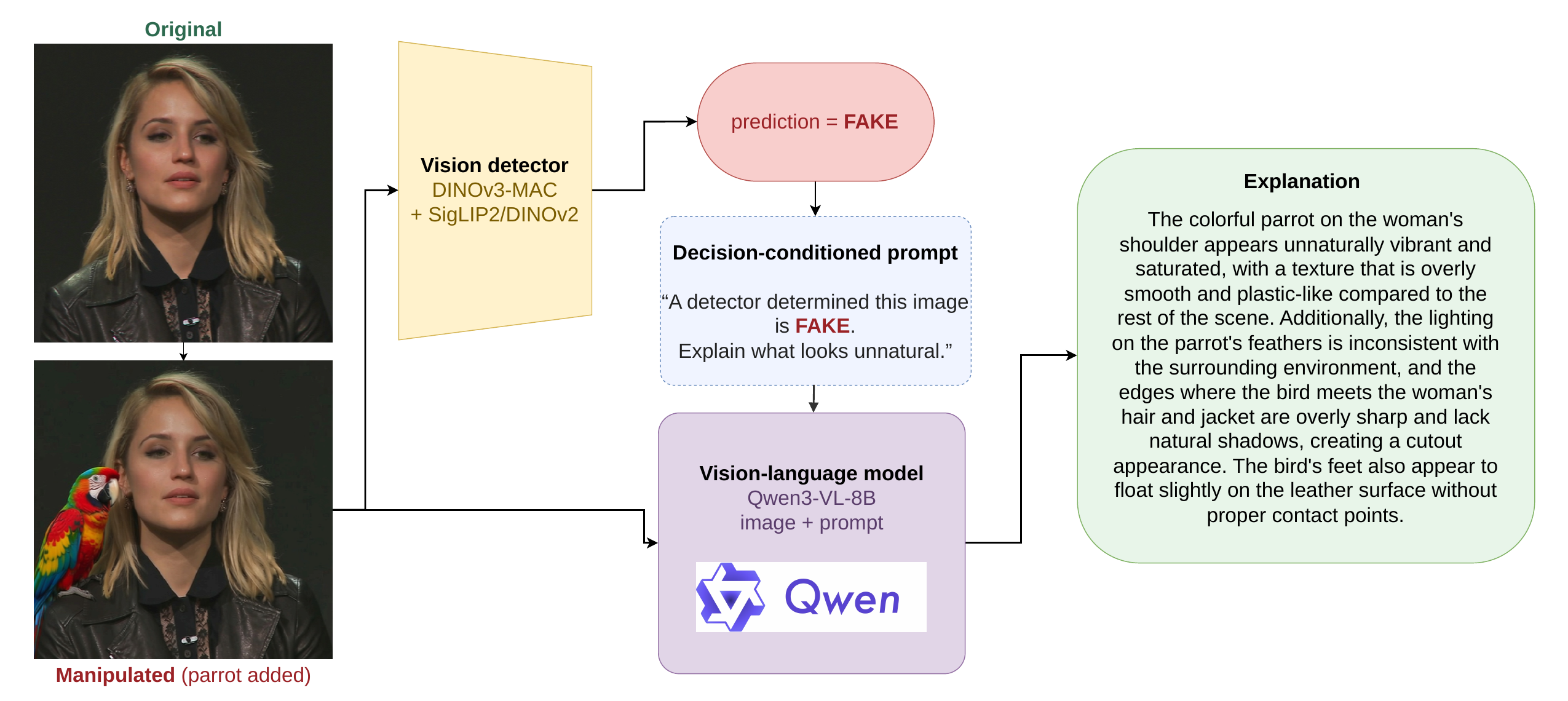}
  \caption{Visualization of our architecture.}
  \Description{A vision detector classifies a manipulated image as fake; a
  GRPO-trained vision-language model then generates a textual explanation of
  the artifacts.}
  \label{fig:teaser}
\end{teaserfigure}


\maketitle


\section{Introduction}

The realism of synthetic imagery has advanced to the point where it is nearly
indistinguishable from authentic content. This is particularly evident for
deepfake imagery, whose low-cost creation is routinely misused for harmful
ends, including misinformation and deliberate attacks on individuals'
reputations. Moving beyond straightforward identity replacement, modern
generative pipelines enable nuanced \emph{person-centric} edits that modify
actions, object interactions, and overall scene semantics while preserving the
subject's identity~\cite{Gupta2025MultiFakeVerse}. As generation methods
advance, detectors still struggle to generalize beyond the generators and
manipulation methods seen during training~\cite{guillaro2025bfree,han2025generaldeepfake},
making reliable identification of manipulated content increasingly demanding.

Despite this growing sophistication of synthetic content, deepfake detection
systems are typically evaluated on classification accuracy alone: a detector
need only output a binary real/fake label, without articulating which regions,
objects, or attributes support that decision. This leaves a critical gap for
real-world deployment, where both experts and non-expert users need to understand
\emph{how} an image was manipulated and which specific changes alter the style,
semantics, or narrative of the scene, especially for localized edits to
actions, interactions, or context, not merely \emph{whether} it was flagged
as synthetic.

The Deepfake Explainability Challenge~\cite{narang2026explainabledeepfakedetectionchallenge}
introduces a new paradigm in which detectors must identify manipulations
and generate grounded natural-language explanations.

In this paper, we present our solution to this challenge: a decoupled
predict-then-explain pipeline that separates foundation-model detection from
decision-conditioned explanation generation. This design provides the
explainer with a predetermined authenticity decision, rather than relying on
the autoregressive output to generate a class-specific narrative before
deciding. Because token-level supervised fine-tuning (SFT) is
only indirectly aligned with the sequence-level metrics used for evaluation,
we follow the recent reinforcement-learning post-training
trend~\cite{ouyang2022instructgpt,deepseekr12025} and further optimize
explanations with GRPO~\cite{shao2024deepseekmath} against the challenge's
semantic-similarity metrics. Our contributions can be summarized as follows:
\begin{itemize}
    \item We introduce a decoupled predict-then-explain pipeline for explainable
    deepfake detection.
    \item The method tackles decision bias and related errors in combined
    decision and explanation generation.
    \item We apply metric-aligned GRPO post-training for both complex and
    simple explanations, improving over SFT alone.
\end{itemize}

\section{Related work}

The detection of AI-generated images has shifted from specialized forensic
systems to vision foundation models
(VFMs)~\cite{lin2024detecting,xu2025fullyaigenerated}. Frozen DINOv3 and
SigLIP2 features with lightweight heads perform strongly on generated-image
and in-the-wild benchmarks~\cite{simplicityprevails2026,vfmwild2025}, while
LoRA adaptation, forensic heads, and complementary VFM ensembles further
improve robustness~\cite{dinov3forensics2026,bombek1ensemble2025,ntire2026}.
These methods achieve strong real-versus-fake classification, but they do not
generally explain the visual evidence behind their decisions.

Recent forensic vision-language models add textual explanations. M2F2-Det
couples a forgery encoder with an LLM so detection features inform both the
fake probability and its explanation~\cite{m2f2det2025}. VIGIL structures
forensic reasoning as a plan-then-examine process over facial
evidence~\cite{vigil2026}. IFDL-VLM~\cite{ifdlvlm2026} decouples detection and localization from
explanation and conditions the VLM on mask-enhanced visual features; we
condition it on binary detector decisions instead. This predict-then-explain
ordering is also more data-efficient than explain-then-predict
training~\cite{jang2021predictfirst}.

GRPO~\cite{shao2024deepseekmath} extends the RL post-training line of
RLHF~\cite{ouyang2022instructgpt} and DeepSeek-R1~\cite{deepseekr12025} by
comparing group-sampled completions without a value model. VIGIL uses
part-aware GRPO for facial forensics~\cite{vigil2026}; we apply metric-aligned
GRPO to the challenge's BERTScore-F1 and SLE rewards.

MultiFakeVerse introduces large-scale person-centric edits that change
actions, interactions, and scene context while preserving
identity~\cite{Gupta2025MultiFakeVerse}. Building on this, XPlainVerse
requires complex and simple natural-language explanations for image-level
detection~\cite{xplainverse2026,narang2026explainabledeepfakedetectionchallenge}.

\section{Methods}
\label{sec:method}

\subsection{Overview}
\label{sec:overview}

Detection and explanation are handled separately in our pipeline
(Figure~\ref{fig:teaser}). Given an input image, a vision-foundation model
(VFM) detector first predicts whether the image is real or fake. The image and
this predetermined decision are then passed to a vision-language model (VLM),
which produces a detailed \emph{complex} explanation of the visible evidence.
For fake images, a second Qwen3-VL-8B summarizes the complex text into a brief
non-expert explanation without using the image. The subsequent subsections
explain the detector, the complex explainer, and the simple explainer.

\subsection{Detection}
\label{sec:pass1}

We employ a Bombek1-style dual-encoder
classifier~\cite{bombek1ensemble2025}. In the first branch, features from
Global SigLIP2-SO400M and DINOv2-Large are concatenated and passed to an MLP
binary classifier. The second branch follows the DINO-MAC architecture used
in the NTIRE 2026 winning solution~\cite{ntire2026}: for each of the final
four DINOv3-Large blocks, mean-pooled patch features are concatenated with
the classification token and four register tokens, then passed to a
layer-specific binary head. All four heads are supervised during training;
during inference, only the final-layer logit is used. In both branches, the
pretrained backbone weights remain frozen; only LoRA adapters in the
attention layers and the binary classification heads are trained. Logits
from each branch are combined using a weight selected on held-out data,
converted into a pseudo-probability, and compared to a threshold to produce
the final real/fake decision. The detector provides only this decision
label; all text generation occurs further down the pipeline.

\subsection{Complex Explanation Generation}
\label{sec:pass2}

The complex explanation is generated by Qwen3-VL-8B-Instruct. During inference, the
image is paired with one of two decision-conditioned prompts. The
\emph{hypothetical-fake} prompt requests visible inconsistencies or synthesis
artifacts, whereas the \emph{hypothetical-real} prompt requests concrete cues
that support authenticity. Both request an evidence-focused paragraph followed by a \texttt{Verdict: real/fake} line, used to reward label consistency during training and stripped before submission. The VLM is adapted via supervised fine-tuning (SFT) with low-rank
adapters on one primary and two hypothetical prompt forms
(Section~\ref{sec:training_corpus}), then
refined with metric-aligned GRPO (Section~\ref{sec:explanation_training}). As a joint baseline, we also train
the same Qwen3-VL-8B with the primary prompt alone so the VLM both decides
and explains; its failure modes motivate the decoupled pipeline
(Section~\ref{sec:results}).

\subsection{Simple Explanation Generation}
\label{sec:simple}

For fake images, a second Qwen3-VL-8B-Instruct model turns the complex
explanation into a short summary in everyday language. This simple explainer
is trained on text-only paired complex and simple references with the same
SFT-then-GRPO recipe, using the competition's simple-explanation metric as the
reward (Section~\ref{sec:explanation_training}). For real images, the complex
explanation is submitted to both output fields, as required by the benchmark.
\section{Experiments}
\label{sec:experiments}

\subsection{Datasets}
\label{sec:dataset}

XPlainVerse, presented for the ACM Multimedia 2026 Deepfake Explainability
Challenge~\cite{xplainverse2026,narang2026explainabledeepfakedetectionchallenge}, builds upon
MultiFakeVerse~\cite{Gupta2025MultiFakeVerse} by incorporating natural-language
explanations.

Each image in the training and validation splits is paired with a binary
real-or-fake label and a detailed \emph{complex} explanation grounded in
visible evidence. Fake images additionally include a concise explanation
written for non-technical users. Table~\ref{tab:xplainverse_split} summarizes
the dataset splits.

\begin{table}[t!]
  \caption{XPlainVerse dataset splits. Test class counts reflect the hidden
  evaluation reference; labels are not included in the public test release.}
  \Description{Train, validation, and test counts of real and fake images
  in XPlainVerse, totaling 760,000 images.}
  \label{tab:xplainverse_split}
  \centering
  \begin{tabular}{lrrr}
    \toprule
    \textbf{Split} & \textbf{Real} & \textbf{Fake} & \textbf{Total} \\
    \midrule
    Training   & 130,000 & 320,000 & 450,000 \\
    Validation &  50,000 &  60,000 & 110,000 \\
    Test       &  90,000 & 110,000 & 200,000 \\
    \midrule
    Total      & 270,000 & 490,000 & 760,000 \\
    \bottomrule
  \end{tabular}
\end{table}

The complex and simple explanation models are trained on the official
XPlainVerse train set (450{,}000 images) and its reference explanations only.
For detection, we held out 1{,}000 images from validation, then combined
official train with the rest of validation and further augmented this set with three external
image-authenticity datasets: OpenFake (1.73M images)~\cite{openfake2025}, the DFBench training
split (436,506)~\cite{dfbench2025}, and the SID-Set training and validation
splits (240,000)~\cite{sida2025}. SID-Set contains equal numbers of real, fully
synthetic, and locally tampered images; we mapped the latter two categories to
fake. These sources
formed the \textbf{2.97M-image mixture} (1.80M fake / 1.16M real) for the SigLIP2--DINOv2 detector.
DINOv3-MAC additionally used 277,643 images from the NTIRE 2026 Robust
AI-Generated Image Detection training set~\cite{ntire2026}, resulting in a
\textbf{3.24M-image mixture} (1.98M fake / 1.26M real).

\subsection{Evaluation Protocols}
\label{sec:evaluation}

\paragraph{Detection}
Labels are binary: 0 indicates a real image, while 1 indicates a generated or
manipulated image. Macro F1 is calculated across both classes using all
200,000 test images.

\paragraph{Reference similarity}
Each prediction requires a detailed complex explanation and a concise simple
explanation, both consistent with the submitted label. Complex explanations
are scored with BERTScore-F1. Simple explanations combine BERTScore-F1 (70\%)
with an SLE simplicity score (30\%), after clipping SLE to $[-1,4]$ and
mapping it to $[0,1]$; we refer to this weighted sum as the simple overall
score. During the competition, the reported reference explanation score was
the average of complex BERTScore-F1 and simple overall.

\paragraph{LLM-based grounding}
The grounding score was omitted during the competition because of its
computational cost. In the post-competition evaluation, a fixed Qwen3.5-4B
judge extracts diagnostic entities and evidence claims from generated and
reference complex explanations. The same judge then measures semantic
coverage in both directions by deciding, for each entity and claim, whether
it is present in the other explanation based on meaning rather than exact
wording. The two directional scores are then combined by their harmonic mean
to produce entity F1 and evidence F1, which are averaged to obtain the
grounding score. The post-competition explanation score is then
$0.4\cdot\text{reference explanation}+0.6\cdot\text{grounding}$, and the
overall score is the mean of this explanation score and detection macro F1.

\subsection{Supervised Training Corpus and Prompt Design}
\label{sec:training_corpus}
We build the explanation corpus from official train and validation
annotations; only the train set is used for training. Each complex sample pairs an image with a
user prompt; the target is the reference text plus
\texttt{Verdict: real/fake}. For the simple tier, each fake complex reference
is paired with its simple summary under a text-only compression prompt.
Complex training uses three templates mixed at about 50\%/25\%/25\%: the
\emph{primary} template asks the model to decide authenticity and justify it
from visible evidence, while \emph{hypothetical-fake} and
\emph{hypothetical-real} provide the ground-truth decision and ask only for
supporting explanation. At inference we use only the hypothetical template
matching the detector prediction, so the explainer does not classify the
image. All templates request multi-region evidence to match reference style
and grounding criteria; the fake-conditioned form asks for structured
technical analysis, and the real-conditioned form uses less technical
language for the typically shorter real references. We keep fake references
with at least three sentences and 80 words and train at a 4:1 fake-to-real
ratio.

\subsection{Detector Training}
\label{sec:detector_training}

Both detector branches are optimized with focal loss
($\gamma{=}2$, $\alpha{=}0.25$) using LoRA
adapters ($r{=}32$, $\alpha{=}64$) on frozen encoders. The SigLIP2--DINOv2
model is warm-started from an XPlainVerse-trained checkpoint and fine-tuned for
five epochs at $392{\times}392$ on the 2.97M-image mixture, with
compression and resize augmentations. DINOv3-MAC is trained from scratch for
two epochs at $512{\times}512$ on the 3.24M-image mixture. We select the best
checkpoint for each branch based on macro F1 on a stratified 1,000-image
holdout set, and use the same holdout to tune the fusion weights and decision
threshold, arriving at 0.6/0.4 and 0.360, respectively.

\subsection{Explanation Model Training}
\label{sec:explanation_training}

The complex explainer and simple compressor are first trained with supervised
fine-tuning (SFT) using LoRA adapters ($r{=}16$, $\alpha{=}32$) on frozen
vision encoders. The complex explainer is Qwen3-VL-8B-Instruct, trained for
one epoch at a learning rate of $1.5{\times}10^{-4}$. The simple compressor
uses the same backbone but processes only text, training for one epoch on
fake-image pairs.

SFT alone was insufficient: training loss kept falling while validation
ROUGE-L plateaued near~$0.35$, reflecting the mismatch between
token-level cross-entropy and the challenge's sequence-level metrics.
We therefore continue with GRPO~\cite{shao2024deepseekmath}: for each
prompt we sample $G$ completions, score them with a scalar reward~$r$,
normalize advantages within the group, and update the policy with a KL
penalty~$\beta$ toward the SFT reference. Simple ($r_{\mathrm{s}}$) and complex ($r_{\mathrm{c}}$) rewards match the
official metrics and are cheap enough for online training:
\begin{align}
  r_{\mathrm{s}}
  &=
  0.7\,\mathrm{BERTScore\text{-}F1}
  +
  0.3\,\mathrm{SLE}_{\mathrm{norm}},
  \label{eq:grpo-rewards}
  \\
  r_{\mathrm{c}}
  &=
  \mathrm{BERTScore\text{-}F1}
  +
  0.3\,r_{\mathrm{fmt}}.
\end{align}
where $\mathrm{BERTScore\text{-}F1}$ uses the leaderboard DeBERTa setup,
$r_{\mathrm{fmt}}$ checks the \texttt{Verdict:} line, and
$\mathrm{SLE}_{\mathrm{norm}}$ follows the official $[0,1]$ mapping.
Both runs use one epoch, $G{=}8$, learning rate $1{\times}10^{-6}$, and
LoRA $(r{=}16,\alpha{=}32)$; complex uses $\beta{=}0.001$ (max length
$400$) and simple $\beta{=}0.04$ (max length~$128$).

\begin{table}[t!]
  \caption{Selected final CodaBench scores~\cite{narang2026explainabledeepfakedetectionchallenge}.
  Complex is Complex BERTScore-F1; Expl.\ is Explanation Score.
  Challenge website: \url{https://explainable-deepfake-detection.github.io/}.}
  \Description{Leaderboard rows for the overall winner, our seventh-place
  submission, and the organizers' Qwen3-VL-8B baseline.}
  \label{tab:final_leaderboard}
  \centering
  \resizebox{0.99\linewidth}{!}{%
  \begin{tabular}{clrrrr}
    \toprule
    \textbf{\#} & \textbf{Team} & \textbf{Overall} & \textbf{Det.\ F1} &
    \textbf{Complex} & \textbf{Expl.} \\
    \midrule
    1 & xz & 0.842 & 0.942 & 0.706 & 0.741 \\
    7 & \textbf{Ours} & 0.791 & 0.860 &
        \textbf{0.712} & 0.722 \\
   32 & Qwen3-VL-8B & 0.643 & 0.634 & 0.665 & 0.651 \\
    \bottomrule
  \end{tabular}%
  }
\end{table}

\section{Results}
\label{sec:results}

\paragraph{Decoupling detection and explanation}
Our primary-prompt VLM baseline showed a strong real bias: on 2{,}000
validation images held out from explanation-model training, 65\% of explanations used a generic real-style opening and
the fake false-negative rate was $44.8\%$. This behavior is consistent with the
semantic-plausibility bias observed in forensic VLMs, which tend to prioritize
coherent scene semantics and surface realism over subtle authenticity
cues~\cite{ifdlvlm2026,tariq2025vlmdeepfake}. The same model is also a weaker
detector than the vision ensemble ($0.772$ vs.\ $0.972$ macro F1 on a separate
1{,}000-image holdout). We therefore let the ensemble decide and ask the VLM
only to explain. Giving the explainer the correct label on those 2{,}000 images
raised complex BERTScore-F1 from $0.698$ to $0.733$.

\paragraph{Effect of GRPO}
To evaluate GRPO's impact, we ran an additional controlled experiment that
directly compared SFT and GRPO complex explanations on the same 10{,}000
explanation-scored test images. Both models received identical images,
detector decisions, decision-conditioned prompts, and decoding settings; only
the explainer adapter differed.
As shown in Table~\ref{tab:grpo_complex}, GRPO raises complex BERTScore-F1 by
$0.026$ and also improves entity F1, evidence F1, and grounding, even
though entity and evidence coverage were never online rewards. Together,
these results indicate that GRPO improves not only complex BERTScore-F1, but
also coverage of the diagnostic entities and evidence claims identified in
the references.

\begin{table}[t]
  \caption{Matched SFT vs.\ GRPO complex-explanation ablation on 10{,}000
  scored test images.}
  \Description{SFT and GRPO adapters compared on complex BERTScore-F1,
  entity F1, evidence F1, and grounding.}
  \label{tab:grpo_complex}
  \centering
  \small
  \begin{tabular}{lcccc}
    \toprule
    \textbf{Adapter} & \textbf{BERTScore-F1} & \textbf{Entity F1} &
    \textbf{Evidence F1} & \textbf{Grounding} \\
    \midrule
    SFT-1655   & 0.686 & 0.362 & 0.259 & 0.310 \\
    GRPO-26400 & \textbf{0.712} & \textbf{0.387} &
                 \textbf{0.296} & \textbf{0.341} \\
    \bottomrule
  \end{tabular}
\end{table}

For simple explanations, swapping only the fake-image compressor on the
hidden test set raised simple overall from $0.677$ to $0.730$, with
detection and complex explanations held fixed.

\paragraph{Final leaderboard}
Table~\ref{tab:final_leaderboard} compares our submission with the
overall winner and the organizers' Qwen3-VL-8B baseline. We ranked
seventh overall. Complex BERTScore-F1 was our strongest component:
$0.712$ ranked first, ahead of the winner's $0.706$. Detection F1
ranked 17th, so detection---not explanation---was the main limitation.

\paragraph{Grounding-aware evaluation}
Finally, we evaluated our final submission with the post-competition protocol
on all 200{,}000 test images. It achieved entity F1 and evidence F1 of
$0.390$ and $0.300$, respectively, a grounding score of $0.345$, an
explanation score of $0.495$, and an overall score of $0.677$
(Table~\ref{tab:paper_results}).

\begin{table}[tb!]
  \caption{Post-competition grounding-aware scores on all 200{,}000 test
  images.}
  \Description{Detection, explanation, grounding, and overall scores of
  our final submission on the full hidden test set.}
  \label{tab:paper_results}
  \centering
  \small
  \begin{tabular}{lr}
    \toprule
    \textbf{Metric} & \textbf{Score} \\
    \midrule
    Detection macro F1            & 0.859709 \\
    Complex BERTScore-F1               & 0.712210 \\
    Simple overall                & 0.727451 \\
    Reference explanation score   & 0.719831 \\
    Entity F1                     & 0.389700 \\
    Evidence F1                   & 0.300414 \\
    Grounding score               & 0.345057 \\
    Explanation score             & 0.494967 \\
    Overall score                 & \textbf{0.677338} \\
    \bottomrule
  \end{tabular}
\end{table}

\section{Discussion}
\label{sec:discussion}

The final leaderboard highlights complex BERTScore-F1 as our strongest component,
ranking first among all submissions. GRPO improved both complex BERTScore-F1 and
the grounding score. Although it used a quick, cost-effective BERTScore-F1
reward instead of entity F1 and evidence F1, its outputs covered more of the
diagnostic entities and evidence claims seen in the references.
Two main limitations remain. First, the reference simple explanations already
favor short motifs such as ``sticker'' and ``doll,'' and metric-aligned GRPO
overuses these patterns in even shorter, more repetitive responses. As a
result, the official simple score rises, mainly via SLE, while linguistic
diversity falls. Second, the predict-then-explain approach propagates detector
errors: while conditioning maintains decision consistency, an incorrect
decision leads the explainer to produce evidence supporting the wrong class.

\section{Conclusion}
\label{sec:conclusion}

These results highlight the value of separating detection from explanation:
specialized vision models determine authenticity, while language models
communicate the supporting evidence for that fixed decision. Metric-aligned
reinforcement learning with GRPO can then improve those explanations using
inexpensive rewards. Because each image has only one reference explanation,
grounding can under-credit plausible alternative evidence; closing this gap
will require grounding-aware judges that are fast enough for online training,
so that GRPO can reward evidence supported by the image rather than BERTScore-F1
alone. Developing such judges, richer reward signals, and comparisons with other
sequence-level methods remains an important direction for future research.

\begin{acks}
Supported in parts by the ARIS Research Programmes P2--0250 (B) and
P2-0214 (B) as well as the ARIS project J2-50065 (A) DeepFake DAD.
\end{acks}
\bibliographystyle{ACM-Reference-Format}
\bibliography{sample-base}


\end{document}